\documentclass{article} 
\usepackage{nips14submit_e,times}
\usepackage{epsfig}
\usepackage{graphicx}
\usepackage{amsmath}
\usepackage{amssymb}
\usepackage{adjustbox}
\usepackage{tabularx}
\usepackage{booktabs}
\usepackage{multirow}
\usepackage{flushend}
\usepackage[font=small,skip=0pt]{caption}
\usepackage{url}

\newcolumntype{m}[1]{>{\centering\arraybackslash}p{#1}}
\usepackage{enumitem}
\setlist{nolistsep}  

\usepackage[pagebackref=true,breaklinks=true,letterpaper=true,colorlinks,bookmarks=false]{hyperref}

\usepackage{blindtext}

\title{SketchParse : Towards Rich Descriptions for Poorly Drawn Sketches using Multi-Task Hierarchical Deep Networks}

\author{
Ravi Kiran Sarvadevabhatla\thanks{This author and co-authors are all affiliated with Video Analytics Lab, Indian Institute of Science, Bangalore, INDIA 560012.} \\
\texttt{ravika@gmail.com} \\
\And
Isht Dwivedi \\
\texttt{isht.dwivedi@gmail.com} \\
\And
Abhijat Biswas \\
\texttt{abhijatbiswas@gmail.com} \\
\And
Sahil Manocha \\
\texttt{sahilmanocha1995@gmail.com} \\
\And
R. Venkatesh Babu \\
\texttt{venky@cds.iisc.ac.in} \\
}

\nipsfinalcopy 

\begin{document}

\maketitle

\begin{abstract}
The ability to semantically interpret hand-drawn line sketches, although very challenging, can pave way for novel applications in multimedia. We propose \textsc{SketchParse}, the first deep-network architecture for fully automatic parsing of freehand object sketches. \textsc{SketchParse} is configured as a two-level fully convolutional network. The first level contains shared layers common to all object categories. The second level contains a number of expert sub-networks. Each expert specializes in parsing sketches from object categories which contain structurally similar parts.  Effectively, the two-level configuration enables our architecture to scale up efficiently as additional categories are added. We introduce a router layer which  (i) relays sketch features from shared layers to the correct expert (ii) eliminates the need to manually specify object category during inference. To bypass laborious part-level annotation, we sketchify photos from semantic object-part image datasets and use them for training. Our architecture also incorporates object pose prediction as a novel auxiliary task which boosts overall performance while providing supplementary information regarding the sketch. We demonstrate \textsc{SketchParse}'s abilities (i) on two challenging large-scale sketch datasets (ii) in parsing unseen, semantically related object categories (iii) in improving fine-grained sketch-based image retrieval. As a novel application, we also outline how \textsc{SketchParse}'s output can be used to generate caption-style descriptions for hand-drawn sketches.
\end{abstract}

\section{Introduction}
\label{sec:intro}

Hand-drawn line sketches have long been employed to communicate ideas in a minimal yet understandable manner. In this paper, we explore the problem of parsing sketched objects, i.e. given a freehand line sketch of an object, determine its salient  attributes (e.g. category, semantic parts, pose). The ability to understand sketches in terms of local (e.g. parts) and global attributes (e.g. pose) can drive novel applications such as sketch captioning, storyboard animation~\cite{kazi2014draco} and automatic drawing assessment apps for art teachers. The onset of deep network era has resulted in architectures which can impressively recognize object sketches at a coarse (category) level~\cite{Sarvadevabhatla:2016:EMR:2964284.2967220,seddati2015deepsketch,yang2015deep}. Paralleling the advances in parsing of photographic objects~\cite{hariharan2015hypercolumns,Liang_2016_CVPR,Xia2016} and scenes~\cite{chen14semantic,noh2015learning,Dai_2016_CVPR}, the time is ripe for understanding sketches too at a fine-grained level~\cite{shoe2016,yi2014bmvc}. 

A number of unique challenges need to be addressed for semantic sketch parsing. Unlike richly detailed color photos, line sketches are binary (black and white) and sparsely detailed. Sketches exhibit a large amount of appearance variations induced by the range of drawing skills among general public. The resulting distortions in object depiction pose a challenge to parsing approaches. In many instances, the sketch is not drawn with a `closed' object boundary, complicating annotation, part-segmentation and pose estimation. Given all these challenges, it is no surprise that only a handful of works exist for sketch  parsing~\cite{SketchSegmentationLabeling:2014,Schneider:2016:ESS:2965650.2898351}. However, even these approaches have their own share of drawbacks (Section \ref{sec:relatedwork}). 

\begin{figure}[!tbp]
\centering
 \includegraphics[width=\linewidth]{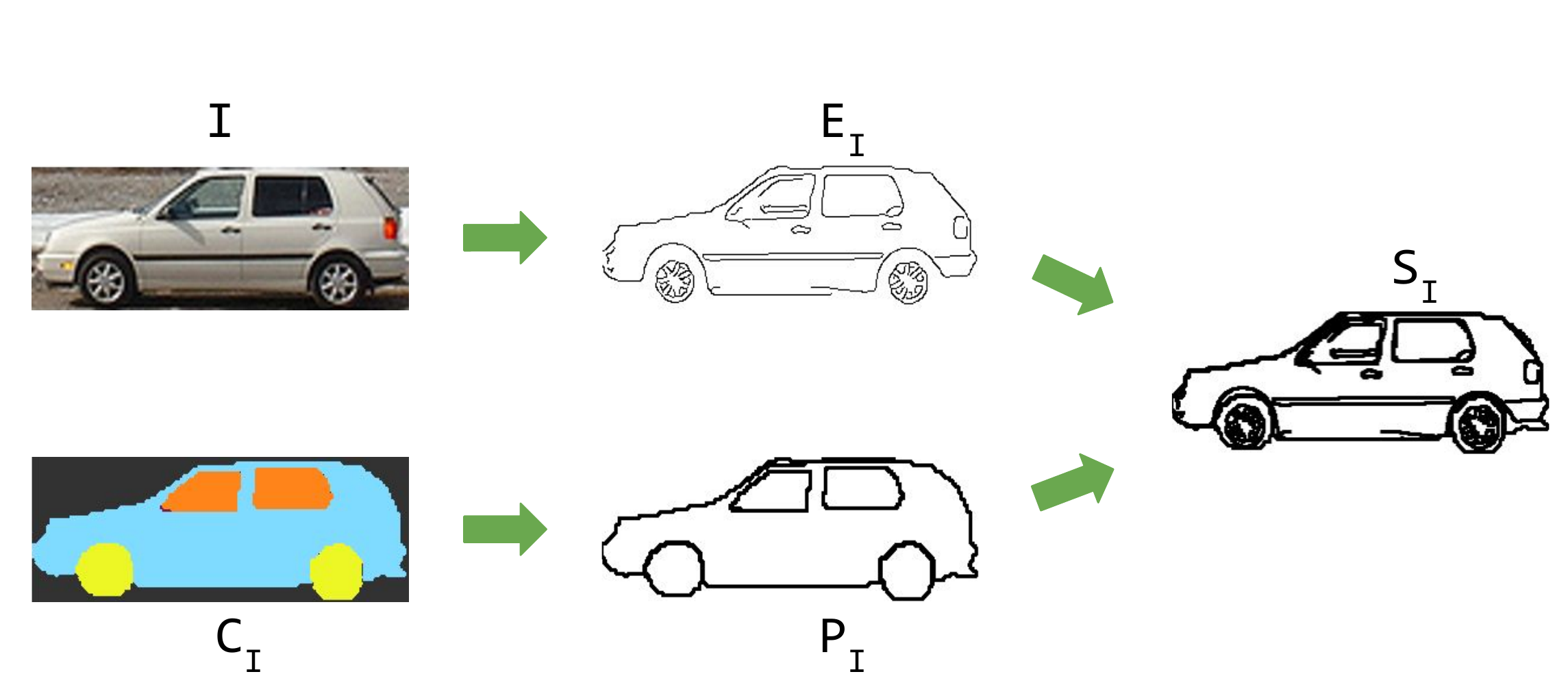}
 \caption{An illustration of our sketchification procedure (Section \ref{sec:getsketchified}). The edge image $E_I$, corresponding to the input photo $I$, is merged with the part and object contours $P_I$ derived from ground-truth labeling $C_I$, to obtain the final sketchified image $S_I$. We train \textsc{SketchParse} using $S_I$ instances as inputs and $C_I$ instances as the corresponding part-labelings.}
\label{fig:sketchification}
\end{figure}

To address these issues, we propose a novel architecture called \textsc{SketchParse} for fully automatic sketch object parsing. In our approach, we make three major design decisions:

\textit{Design Decision \#1 (Data):} To bypass burdensome  part-level sketch annotation, we leverage photo image datasets containing part-level annotations of objects~\cite{chen_cvpr14}. Suppose $I$ is an object image and $C_I$ is the corresponding part-level annotation. We subject $I$ to a sketchification procedure (Section \ref{sec:getsketchified}) and obtain $S_I$. Thus, our training data consists of the sketchified image and corresponding part-level annotation pairs ($\{S_I,C_I\}$) for each category (Figure \ref{fig:sketchification}). 

\textit{Design Decision \#2 (Model):} Many structurally similar object categories tend to have common parts. For instance, `wings' and `tail' are common to both birds and airplanes. To exploit such shared semantic parts, we design our model as a two-level network of disjoint experts (see Figure \ref{fig:sketchlab-overview}). The first level contains shared layers common to all object categories. The second level contains a number of experts (sub-networks). Each expert is configured for parsing sketches from a super-category set comprising of categories with structurally similar parts\footnote{\label{note1}For example, categories \texttt{cat,dog,sheep} comprise the super-category \textit{Small Animals}. }. Instead of training from scratch, we instantiate our model using two disjoint groups of pre-trained layers from a scene parsing net (Section \ref{sec:meth-dl-splitting}). We perform training using the sketchified data (Section \ref{sec:training}) mentioned above. At test time, the input sketch is first processed by the shared layers to obtain intermediate features. In parallel, the sketch is also provided to a super-category sketch classifier. The label output of the classifier is used to automatically route the intermediate features to the appropriate super-category expert for final output i.e. part-level segmentation (Section \ref{sec:inference}).

\textit{Design Decision \#3 (Auxiliary Tasks):} A popular paradigm to improve performance of the main task is to have additional yet related auxiliary targets in a multi-task setting~\cite{ranjan2016hyperface,abdulnabi2015multi,lapin2014scalable,mahasseni2013latent}. Motivated by this observation, we configure each expert network for the novel auxiliary task of 2-D pose estimation.

At first glance, our approach seems infeasible. After all, sketchified training images resemble actual sketches only in terms of stroke density (see Figure \ref{fig:sketchification}). They seem to lack the fluidity and unstructured feel of hand-drawn sketches. Moreover, \textsc{SketchParse}'s base model~\cite{chen2016deeplab}, originally designed for \underline{photo} \textit{scene} segmentation, seems an unlikely candidate for enabling transfer-learning based \underline{sketch} \textit{object} segmentation. Yet, as we shall see, our design choices result in an architecture which is able to successfully accomplish sketch parsing across multiple categories and sketch datasets. 

\noindent \textbf{Contributions:}

\begin{itemize}
    \item We propose \textsc{SketchParse} -- the first deep hierarchical network for fully automatic parsing of hand-drawn object sketches (Section \ref{sec:methodology}). Our architecture includes object pose estimation as a novel auxiliary task. 
    \item We provide the largest dataset of part-annotated object sketches across multiple categories and multiple sketch datasets. We also provide 2-D pose annotations for these sketches.
    \item We demonstrate \textsc{SketchParse}'s abilities on two challenging large-scale sketch object datasets (Section \ref{sec:inference}), on unseen semantically related categories, (Section \ref{sec:experiments-relatedcategories}) and for improving fine-grained sketch-based image retrieval (Section \ref{sec:experiments-fgr}).
    \item We outline how \textsc{SketchParse}'s output can form the basis for novel applications such as automatic sketch description (Section \ref{sec:experiments-fgr}).
\end{itemize}

Please visit \url{https://github.com/val-iisc/sketch-parse} for pre-trained models, code and resources related to the work presented in this paper.

\section{Related Work}
\label{sec:relatedwork}

\textbf{Semantic Parsing (Photos):} Existing deep-learning approaches for semantic parsing of photos can be categorized into two groups. The first group consists of approaches for scene-level semantic parsing (i.e. output an object label for each pixel in the scene)~\cite{chen2016deeplab,noh2015learning,Dai_2016_CVPR}. The second group of approaches attempt semantic parsing of objects (i.e. output a part label for each object pixel)~\cite{hariharan2015hypercolumns,Liang_2016_CVPR,Xia2016}. Compared to our choice (of a scene parsing net), this latter group of object-parsing approaches seemingly appear better candidates for the base architecture. However, they pose specific difficulties for adoption. For instance, the hypercolumns approach of Hariharan et al.~\cite{hariharan2015hypercolumns} requires training separate part classifiers for each class. Also, the evaluation is confined to a small number of classes (animals and human beings). The approach of Liang et al.~\cite{Liang_2016_CVPR} consists of a complex multi-stage hybrid CNN-RNN architecture evaluated on only two animal categories (\texttt{horse} and \texttt{cow}). The approach of Xia et al.~\cite{Xia2016} fuses object part score evidence from scene, object and part level to obtain impressive results for two categories (human, large animals). However, it is not clear how their method can be adopted for our purpose.     

\noindent \textbf{Semantic Object Parsing (Sketches):} Only a handful of works exist for sketch parsing~\cite{SketchSegmentationLabeling:2014,Schneider:2016:ESS:2965650.2898351}.  Existing approaches require a part-annotated dataset of sketch objects. Obtaining such part-level annotations is very laborious and cumbersome. Moreover, one-third of a dataset~\cite{SketchSegmentationLabeling:2014} evaluated by these approaches consists of sketches drawn by professional artists. Given the artists' relatively better drawing skills, incorporating such sketches artificially reduces the complexity of the problem. On the architectural front, the approaches involve tweaking of multiple parameters in a hand-crafted segmentation pipeline. Existing approaches label individual sketch strokes as parts. This requires strokes within part interiors to be necessarily labelled, which can result in peculiar segmentation errors~\cite{Schneider:2016:ESS:2965650.2898351}. Our method, in contrast, labels object regions as parts. In many instances, the object region boundary consists of non-stroke pixels. Therefore, it is not possible to directly compare with existing approaches. Unlike our category-scalable and fully automatic approach, these methods assume object category is known and train a separate model per category (E.g. \texttt{dog} and \texttt{cat} require separate models). Existing implementations of these approaches also have prohibitive inference time -- parsing an object sketch takes anywhere from $2$ minutes~\cite{Schneider:2016:ESS:2965650.2898351} to $40$ minutes~\cite{SketchSegmentationLabeling:2014}, rendering them unsuitable for interactive sketch-based applications. In contrast, our model's inference time is fraction of a \textit{second}. Also, our scale of evaluation is significantly larger. For example, Schneider et al.'s method~\cite{Schneider:2016:ESS:2965650.2898351} is evaluated on 5 test sketches per category. Our model is evaluated on $100$ sketches per category. Finally, none of the previous approaches exploit the hierarchical category-level groupings which arise naturally from structural similarities~\cite{zhao2011large}. This renders them prone to drop in performance as additional categories (and their parts) are added. 

\noindent \textbf{Sketch Recognition:} The initial performance of handcrafted feature-based approaches~\cite{Schneider2014} for sketch recognition has been surpassed in recent times by deep CNN architectures~\cite{yang2015deep,seddati2015deepsketch}. The sketch router classifier in our architecture is a modified version of Yang et al.'s Sketch-a-Net~\cite{yang2015deep}. While the works mentioned above use sketches, Zhang et al.~\cite{ZhangFreehand} use sketchified photos for training the CNN classifier which outputs class labels. We too use sketchified photos for training. However, the task in our case is parsing and not classification.

\noindent \textbf{Class-hierarchical CNNs:} Our idea of having an initial coarse-category net which routes the input to finer-category experts can be found in some recent works as well~\cite{Ahmed2016,yan2015hd}, albeit for object classification. In these works, coarse-category net is intimately tied to the main task (viz. classification). In our case, the coarse-category CNN classifier serves a secondary role, helping to route the output of a parallel, shared sub-network to the finer-category parsing experts. Also, unlike above works, the task of our secondary net (classification) is different from the task of experts (segmentation).

\noindent \textbf{Domain Adaptation/Transfer Learning:} Our approach can be viewed as belonging to the category of domain adaptation techniques~\cite{saenko2010adapting,patel2015visual,bergamo2010exploiting}. These techniques have proven to be successful for various problems, including image parsing~\cite{ros2016synthia,hong2015TransferNet,van2015transfer}. However, unlike most approaches wherein image modality does not change, our domain-adaptation scenario is characterized by extreme modality-level variation between source (image) and target (freehand sketch). This drastically reduces the quantity and quality of data available for transfer learning, making our task more challenging.

\noindent \textbf{Multi-task networks:} The effectiveness of addressing auxiliary tasks in tandem with the main task has been shown for several challenging problems in vision~\cite{ranjan2016hyperface,abdulnabi2015multi,lapin2014scalable,mahasseni2013latent}. In particular, object classification~\cite{DBLP:journals/corr/NekrasovJC16}, detection~\cite{Dai_2016_CVPR}, geometric context~\cite{vezhnevets2010towards}, saliency~\cite{li2016deepsaliency} and adversarial loss~\cite{luc2016semantic} have been utilized as auxiliary tasks in deep network-based approaches for semantic parsing. The auxiliary task we employ -- object viewpoint estimation -- has been used in a multi-task setting but for object classification~\cite{DBLP:journals/corr/ZhaoI16a,Elhoseiny:2016:CAS:3045390.3045485}. For instance, Wong et al.~\cite{wong2017segicp} project the output of semantic scene segmentation onto a depth map and use a category-aware 3D model approach to enable 3-D pose-based grasping in robots. Zhao and Itti~\cite{DBLP:journals/corr/ZhaoI16a} utilize 3-D pose information from toy models of $10$ categories as a target auxiliary task to improve object classification. Elhoseiny et al.~\cite{Elhoseiny:2016:CAS:3045390.3045485} also introduce pose-estimation as an auxiliary task along with classification. To the best of our knowledge, we are the first ones to design a custom pose estimation architecture to assist semantic parsing.

\section{Data Preparation}
\label{sec:datapreparation}

We first summarize salient details of two semantic object-part photo datasets. 

\subsection{Object photo datasets}
\label{sec:datapreparation-objphotodatasets}

\noindent \textbf{PASCAL-Parts:} This $10{,}103$ image dataset~\cite{wang2015joint} provides semantic part segmentation of objects from the $20$ object categories from the PASCAL VOC2010 dataset. We select $11$ categories (\texttt{aeroplane, bicycle, bus, car, cat, cow, dog, flying bird, horse, motorcycle, sheep}) for our experiments. To obtain the cropped object images, we used object bounding box annotations from PASCAL-parts. 

\noindent \textbf{CORE:} The Cross-category Object REcognition (CORE) dataset~\cite{farhadi2010attribute} contains segmentation and attribute information for objects in $2800$ images distributed across $28$ categories of vehicles and animals. We select $8$ categories from CORE dataset based on their semantic similarity with PASCAL-part categories (e.g. CORE category \texttt{crow} is selected since it is semantically similar to the PASCAL-part category \texttt{bird}).

To enable estimation of object pose as an auxiliary objective, we annotated all the images from PASCAL-Parts and CORE datasets with 2-D pose information based on the object's orientation with respect to the viewing plane. Specifically, each image is labeled with one of the cardinal (`North',`East', `West',`South') and intercardinal directions (`NE','NW',`SE',`SW') ~\cite{wikicardinal}. We plan to release these pose annotations publicly for the benefit of multimedia community.

Next, we shall describe the procedure for obtaining sketchified versions of photos sourced from these datasets.

\subsection{Obtaining sketchified images}
\label{sec:getsketchified}

Suppose $I$ is an object image. As the first step, we use a Canny edge detector tuned to produce only the most prominent object edges. Visually, we found the resulting image $E_I$ to contain edge structures which perceptually resemble human sketch strokes compared to those produced by alternatives such as Sketch Tokens~\cite{lim2013sketch} and SCG~\cite{xiaofeng2012discriminatively}. We augment $E_I$ with part contours and object contours from the part-annotation data of $I$ and perform morphological dilation to thicken edge segments (using a square structured element of side $3$) to obtain the sketchified image $S_I$ (see Figure \ref{fig:sketchification}). To augment data available for training the segmentation model, we apply a series of rotations ($\pm10$, $\pm20$, $\pm30$ degrees) and mirroring about the vertical axis to $S_I$. Overall, this procedure results in $14$ augmented images per each original, sketchified image. To ensure a good coverage of parts and eliminate inconsistent labelings, we manually curated $1532$ object images from PASCAL-Parts and CORE datasets. Given the varying semantic granularity of part labels across these datasets, we manually curated the parts considered for each category~\cite{wang2015joint}. Finally, we obtain the training dataset consisting of $1532 \times 14 = 21{,}448$ paired sketchified images and corresponding part-level annotations, distributed across $11$ object categories.

We evaluate our model's performance on freehand line sketches from two large-scale datasets. We describe these datasets and associated data preparation procedures next. 

\subsection{Sketch datasets and augmentation}
\label{sec:sketchdatasets}

\noindent \textbf{TU-Berlin:} The TU-Berlin sketch database~\cite{eitz2012humans} contains $20{,}000$ hand drawn sketches spanning $250$ common object categories, with $80$ sketches per object. For this dataset, only the category name was provided to the sketchers during the drawing phase.  

\noindent \textbf{Sketchy:} The Sketchy database~\cite{Sangkloy:2016:SDL:2897824.2925954} contains $75{,}471$ sketches spanning $125$ object categories, with $500-700$ sketches per category. To collect this dataset, photo images of objects were initially shown to human subjects. After a gap of $2$ seconds, the image was replaced by a gray screen and  subjects were asked to sketch the object from memory. Compared to the draw-from-category-name-only approach employed for TU-Berlin dataset~\cite{eitz2012humans}, this memory-based approach provides a larger variety in terms of multiple viewpoints and object detail in sketches. On an average, each object photo is typically associated with $5-8$ different sketches. 

From both the datasets, we use sketches from only those object categories which overlap with the $11$ categories from PASCAL-Parts mentioned in Section \ref{sec:datapreparation-objphotodatasets}. For augmentation, we first apply morphological dilation (using a square structured element of side $3$) to each sketch. This operation helps minimize the impact of loss in stroke continuity when the sketch is processed by deeper layers of the network. Subsequently, we apply a series of rotations ($\pm10$, $\pm20$, $\pm30$ degrees) and mirroring about the vertical axis on the dilated sketch. This produces $14$ augmented variants per original sketch for use during training.

\section{Our model (\textsc{SketchParse})}
\label{sec:methodology}

\begin{figure*}[!htbp]
\centering
\includegraphics[width=\textwidth,keepaspectratio]{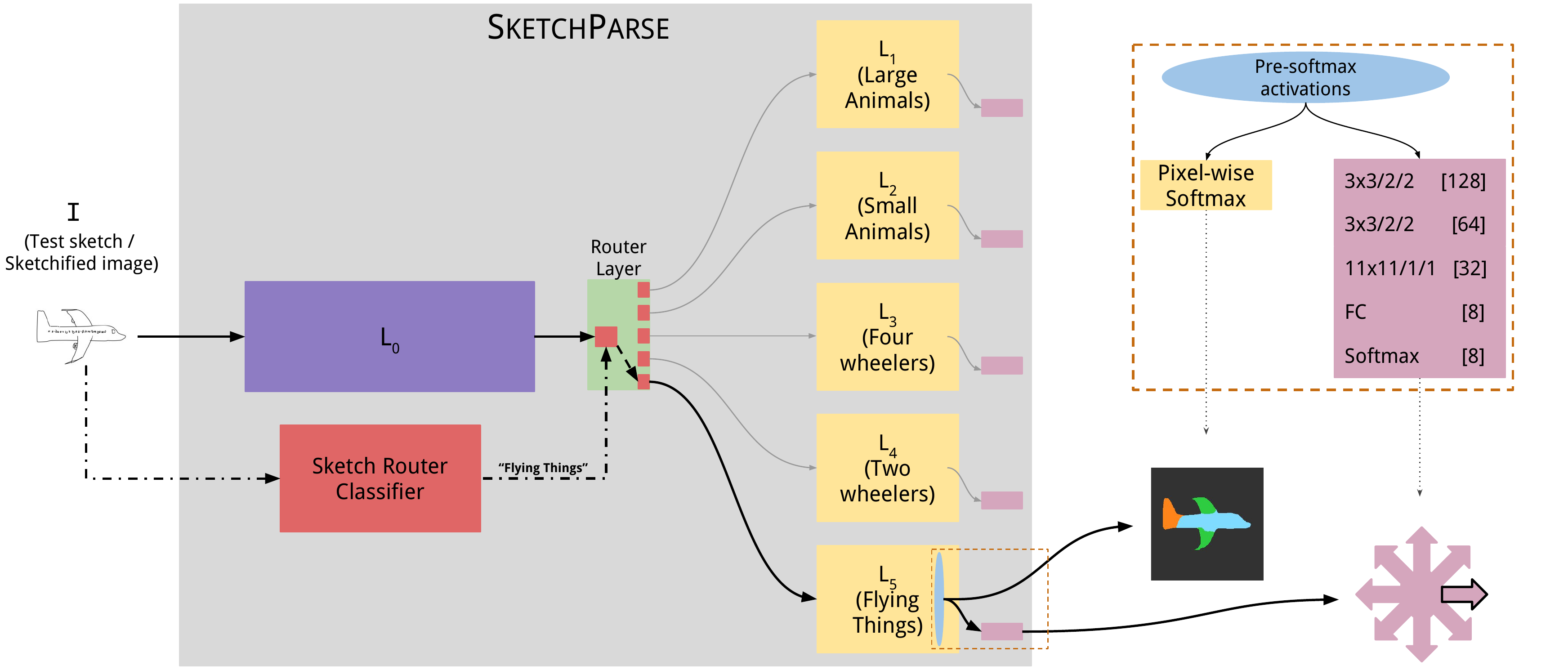}
\caption{The first level $L_0$ of \textsc{SketchParse} (shaded purple) is instantiated with shallower layers of a scene parsing net. The second level consists of $K$ expert super-category nets $L_1,L_2,\ldots L_K$ (shaded yellow) and is instantiated with deeper layers of the scene parsing net. Given the test sketch $I$, the Router Layer (shaded green) relays intermediate features produced by shared layers $L_0$ to the target expert ($L_5$). The pre-softmax activations (blue oval) generated in $L_5$ are used to obtain the part parsing. These activations are also used by our novel pose estimation auxiliary net (shaded light-pink). The architecture of the pose net can be seen within the brown dotted line box in top-right. Within the pose net, convolutional layers are specified in the format dimensions/\hspace{0.25mm}stride/\hspace{0.25mm}dilation\hspace{0.5mm}[number of filters]. FC = Fully-Connected layer. The dash-dotted line connected to the router classifier is used to indicate that Router Layer is utilized only during inference.}
\label{fig:sketchlab-overview}
\end{figure*}

\subsection{Instantiating \textsc{SketchParse} levels}
\label{sec:meth-dl-splitting}

We design a two-level deep network architecture for \textsc{SketchParse} (see Figure \ref{fig:sketchlab-overview}). The first level is intended to capture category-agnostic low-level information and contains shared layers $L_0$ common to all $N$ object categories. We instantiate first-level layers with the shallower, initial layers from a scene parsing net~\cite{chen2016deeplab}. Specifically, we use multi-scale version of ResNet-101~\cite{he2016deep} variant of DeepLab~\cite{chen2016deeplab}, a well-known architecture designed for semantic scene parsing. In this context, we wish to emphasize that our design is general and can accommodate any fully convolutional scene parsing network. 

The categories are grouped into $K$ smaller $(< N)$, disjoint super-category subsets using meronym (i.e. part-of relation)-based similarities between objects~\cite{theobald03}. To obtain super-categories, we start with given categories and cluster them based on the fraction of common meronyms (part-names). For example, `flying bird' and `airplane' have the largest common set of part-names between themselves compared to any other categories and therefore form a single cluster. This procedure can also be used when a new category $C$ is introduced -- we assign $C$ to the super-category with which it shares largest common set of part-names. 

The second level consists of $K$ expert sub-networks $L_1,L_2,\ldots L_K$, each of which is specialized for parsing sketches associated with a super-category. We initialize these $K$ experts using the deeper, latter layers from the scene parsing model. Suppose the total number of parts across all object categories within the $i$-th super-category is $n_i, 1 \leqslant i \leqslant K$. We modify the final layer for each  expert network $L_i$ such that it outputs $n_i$ part-label predictions. In our current version of the architecture, $K=5$ and $N=11$. We performed ablative experiments to determine optimal location for splitting the layers of semantic scene parsing net into two groups. Based on these experiments, we use all the layers up to the \textit{res5b} block as shared layers. 

\subsection{Router Layer}
\label{sec:meth-sketchrouter}

From the above description (Section \ref{sec:meth-dl-splitting}), \textsc{SketchParse}'s design so far consists of a single shared sub-network and $K$ expert nets. We require a mechanism for routing the intermediate features produced by shared layers to the target expert network. In addition, we require a mechanism for backpropagating error gradients from the $K$ expert nets and update the weights of the shared layer sub-network during training. To meet such requirements, we design a Router Layer (shaded green in Figure \ref{fig:sketchlab-overview}). During the training phase, routing of features from the shared layer is dictated by ground-truth category and by extension, the super-category it is associated with. A branch index array is maintained for each training mini-batch. Since the ground-truth super-category for each training example is available, creation of branch index array requires only knowledge of the mini-batch label composition. The array entries are referenced during backward propagation to (a) recombine the gradient tensors in the same order as that of the mini-batch and (b) route error gradient from the appropriate branch to the shared layers during backpropagation.

To accomplish routing during test time, we use a $K$-way classifier (shaded red in Figure \ref{fig:sketchlab-overview}) whose output label corresponds to one of the $K$ expert networks $L_1,L_2,\ldots L_K$. In this regard, we experimented with a variety of deep CNN architectures. We examined previous custom architectures for sketch recognition by Seddati et al.~\cite{seddati2015deepsketch} and Yang et al.~\cite{yang2015deep}. In addition to custom nets, we explored architectures which involved fine-tuning image object classification networks such as AlexNet and GoogleNet. Table \ref{arch:comp} lists some of the architectures explored along with their classification performance. We briefly describe the architectures next.

\begin{itemize}
    \item \textbf{[\textsc{SketchParse}-pool5][GAP][FC64][DO 0.5][FC5], using sketchified images} - We first train \textsc{SketchParse} and then take global average pooled output after pool5. All the layers of \textsc{SketchParse} are frozen, only the fully connected layers are learnt. 0.5 dropout is used for the fully connected layer.
    \item \textbf{[\textsc{SketchParse}-pool5][FC64][DO 0.5][FC64][DO 0.5][FC5], using sketchified images} - This experiment is similar to the above experiment. Here we use 3 fully connected layers after the global average pool output of pool5. 0.5 dropout is used for the fully connected layers.
    \item \textbf{ Google net [GAP][FC5], using sketches} - The last fully connected layer of the Google net is removed and it is replaced by a global average pool layer and a fully connected layer for classification. This model is then fine-tuned for sketch classification. 
    \item \textbf{ Custom net (used), using sketches} - This is the custom net, described in Table \ref{arch}. It was found that on increasing the number of layers in all Conv layers of this architecture improves performance.
    \item \textbf{Alex net [GAP][FC5], using sketches} - In this experiment, we replaced the last 3 fully connected layers of Alex net with a global average pool layer and a fully connected layer for classification. This new architecture was then fine-tuned. 
\end{itemize}

\begin{table*}[!htbp]
\footnotesize
\centering
\begin{tabular}{ccc}
\hline
Architecture & training data &performance \\ 
\hline
 {[}\textsc{SketchParse}-pool5]{[}GAP]{[}FC64]{[}DO 0.5]{[}FC5]& sketchified images & $67.0$ \\
  {[}\textsc{SketchParse}-pool5]{[}FC64]{[}DO 0.5]{[}FC64]{[}DO 0.5]{[}FC5]& sketchified images & $64.0$ \\
 Google net {[}GAP][FC5] & sketch images & $90.0$ \\
 Custom net(used) & sketch images & $\textbf{91.3}$\\  
 Alex net {[}GAP][FC5] & sketch images & $89.8$  \\ 
 \hline
\end{tabular}
\caption{Some of the architectures explored to find the best classifier and their performance numbers.}
\label{arch:comp}
\end{table*}

We found that a customized version of Yang et al.'s SketchCNN architecture, with a normal (non-sketchified) input, provided the best performance. Specifically, we found that  (1) systematically increasing the number of filters in all the conv layers and (2) using a larger size kernel in last layer, results in better performance. We use a very high dropout rate of $70$\% in our network to combat over-fitting and to compensate for lower amount of training data. Table \ref{arch} depicts the architecture of our router classifier. 
 
We also wish to point out that our initial attempts involved training custom CNNs solely on sketchified images or their deep feature variants. However, the classification performance was subpar. Therefore, we resorted to training the classifier using actual sketches.

\begin{table*}[!tbp]
\footnotesize
\centering
\begin{tabular}{ccccc}
\hline
index & Type & Filter  size& no. of filters & stride \\ 
\hline\hline
 1 & Conv & 15x15 & 64 & 3\\ 
 2 & ReLu & - & - & -\\ 
 3 & Maxpool & 3x3 & - & 2\\ 
 4 & Conv & 5x5 & 128 & 1\\ 
 5 & ReLu & - & - & -\\ 
 6 & Maxpool & 3x3 & - & 2\\ 
 7 & Conv & 3x3 & 256 & 1\\ 
 8 & ReLu & - & - & -\\ 
 9 & Conv & 3x3 & 256 & 1\\ 
 10 & ReLu & - & - & -\\ 
 11 & Conv & 3x3 & 256 & 1\\ 
 12 & ReLu & - & - & -\\ 
 13 & Maxpool & 3x3 & - & 2\\ 
 14 & Conv & 1x1 & 512 & -\\ 
 15 & ReLu & - & - & -\\ 
 16 & Dropout(0.7) & - & - & -\\ 
 17 & Conv & 1x1 & 5 & -\\ \hline
\end{tabular}
\caption{The architecture of the sketch router classifier used by \textsc{SketchParse}.}
\label{arch}
\end{table*}

\subsubsection{Confusion Matrix}

A highly accurate sketch classifier is a crucial requirement for proper routing and overall performance of our approach. Table \ref{tab:conf} shows the confusion matrix of the classifier used by our model on the test set.  We note that most of the misclassification occurs due to confusion between \textit{Small Animal} and \textit{Large Animal} super-categories because these sketches tend to be very similar. 

\begin{table*}[!htbp]

\footnotesize
\centering
\begin{tabular}{|c|c|c|c|c|c|}
\hline
 & Large Anim. & Small Anim. & 4 wheel. & 2 wheel. &  Flying. \\ 
\hline\hline
 Large Anim. & $80.05$ & $9.40$ & $0$ & $0$ & $1$ \\ \hline
Small Anim.  & $18.9$ & $88.35$ & $0.15$ & $0.15$ & $3.15$ \\ \hline
4 wheel.  & $0.08$ & $0.25$ & $98.80$ & $0$ & $2.0$ \\ \hline
2 wheel.   & $0.4$ & $0.89$ & $0.5$ & $99.15$ & $0.4$ \\ \hline
Flying.   & $0.5$ & $0.94$ & $0.50$ & $0.65$ & $93.25$ \\ \hline
\end{tabular}
\caption{The confusion matrix of the classifier used by our model on the test set.}
\label{tab:conf}
\end{table*}

\subsection{Auxiliary (Pose Estimation) Task Network}
\label{sec:meth-auxtask}

The architecture for estimating the 2-D pose of the sketch (shaded pink and shown within the top-right brown dotted line box in Figure \ref{fig:sketchlab-overview}) is motivated by the following observations:

\textit{First,} the part-level parsing of the object typically provides clues regarding object pose. For example, if we view the panel for \textit{Two Wheelers} in Figure  \ref{fig:qualitative-sketches}, it is evident that the relative location and spatial extent of `handlebar' part for a \texttt{bicycle} is a good indicator of pose. Therefore, to enable pose estimation from part-level information, the input to the pose net is the tensor of pre-softmax pixelwise activations\footnote{Shown as a blue oval in Figure \ref{fig:sketchlab-overview}.} generated within the expert part-parsing network. 
To capture the large variation in part appearances, locations and combinations thereof, the first two layers in the pose network contain dilated convolutional filters ~\cite{yu2015multi}, each having rate $r=2$ and stride $s=2$ with kernel width $k=3$. Each convolutional layer is followed by a ReLU non-linearity~\cite{nair2010rectified}. 

\textit{Second,} 2-D pose is a global attribute of an object sketch. Therefore, to provide the network with sufficient global context, we configure the last convolutional layer with $r=s=1$ and $k=11$, effectively learning a large spatial template filter. The part combinations captured by initial layers also mitigate the need to learn many such templates -- we use $32$ in our implementation. The resulting template-based feature representations are combined via a fully-connected layer and fed to a $8$-way softmax layer which outputs 2-D pose labels corresponding to cardinal and intercardinal directions. Note also that each super-category expert has its own pose estimation auxiliary net.   

Having described the overall framework for \textsc{SketchParse}, we next describe major implementation details of our training and inference procedure.

\section{Implementation Details}
\label{sec:experiments-setup}

\subsection{Training}
\label{sec:training}

\noindent \textbf{\textsc{SketchParse}}: Before commencing \textsc{SketchParse} training, the initial learning rate for all but the final convolutional layers is set to $5 \times 10^{-4}$. The rate for the final convolutional layer in  each sub-networks is set to $5 \times 10^{-3}$. Batch-norm parameters are kept fixed during the entire training process. The architecture is trained end-to-end using a per-pixel cross-entropy loss as the objective function. For optimization, we employ stochastic gradient descent with a mini-batch size of $1$ sketchified image, momentum of $0.9$ and polynomial weight decay policy. We stop training after $20000$ iterations. The training takes $3.5$ hours on a NVIDIA Titan X GPU. 

A large variation can exist between part sizes for a given super-category (e.g. number of `tail' pixels is smaller than `body' pixels in \textit{Large Animals}). To accommodate this variation, we use a class-balancing scheme which weighs per-pixel loss differently based on relative presence of corresponding ground-truth part~\cite{eigen2015predicting}. Suppose a pixel's ground-truth part label is $c$. Suppose $c$ is present in $N_c$ training images and suppose the total number of pixels with label $c$ across these images is $p_c$.  We weight the corresponding loss by $\alpha_c = M/f_c$ where $f_c = p_c/N_c$ and $M$ is the median over the set of $f_c$s. In effect, losses for pixels of smaller parts get weighted to a larger extent and vice-versa. 

\noindent \textbf{Pose Auxiliary Net}: The pose auxiliary network is trained end-to-end with the rest of \textsc{SketchParse}, using an $8$-way cross-entropy loss. The learning rate for the pose module is set to $2.5 \times 10^{-2}$. All other settings remain the same as above.

\noindent \textbf{Sketch classifier:} For training the $K$-way sketch router classifier, we randomly sample $40\%$ of sketches per category from TU-Berlin and Sketchy datasets\footnote{This translates to $26$ sketches from TU-Berlin and $170$ sketches from Sketchy for training, $6$ and $30$ sketches for validation.}. We augment data with flip about vertical axis, series of rotations ($\pm4$, $\pm8$, $\pm12$ degrees) and sketch-scale augmentations ($\pm3$\%, $\pm7$\% of image height). For training, the initial learning rate is set to $7 \times 10^{-4}$. The classifier is trained using stochastic gradient descent with a mini-batch size of $600$ sketches, momentum of $0.9$ and a polynomial weight decay policy. We stop training after $20000$ iterations. The training takes $4$ hours on a NVIDIA Titan X GPU. 

\subsection{Inference}
\label{sec:inference}

For evaluation, we used equal number of sketches ($48$) per category from TU-Berlin and Sketchy datasets except for \texttt{bus} category which is present only in TU-Berlin dataset. Thus, we have a total of $(48 \times 2 \times 10) + 48 = 1008$ sketches for testing. For sketch router classifier, we follow the conventional approach~\cite{seddati2015deepsketch} of pooling score outputs corresponding to cropped (four corner crops and one center crop) and white-padded versions of the original sketch and its vertically mirrored version. Overall, the time to obtain part-level segmentation and pose for an input sketch is $0.25$ seconds on average. Thus, \textsc{SketchParse} is an extremely suitable candidate for developing applications which require real-time sketch understanding.

\section{Experiments}
\label{sec:experiments}

To enable quantitative evaluation, we crowdsourced part-level annotations for all the sketches across $11$ categories, in effect creating the largest part-annotated dataset for sketch object parsing. 

\noindent \textbf{Evaluation procedure:} For quantitative evaluation, we adopt the average IOU measure widely reported in photo-based scene and object parsing literature~\cite{long2015fully}. Consider a fixed test sketch. Let the number of unique part labels in the sketch be $n_p$. Let $n_{ij}$ be the number of pixels of part-label $i$ predicted as part-label $j$. Let $t_i = \sum_j n_{ij}$ be the total number of pixels whose ground-truth label is $i$.  We first define part-wise Intersection Over Union for part-label $i$ as \hspace{0.5mm} $pwIOU_{i}$ =  $\frac{n_{ii}}{t_i + {\sum_j n_{ji}} - n_{ii}}$. Then, we define sketch-average IOU (sIOU) = $ \sum_i \frac{pwIOU_{i}}{n_p}$. For a given category, we compute sIOU for each of its sketches individually and average the resulting values to obtain the category's average IOU (aIOU) score. 

\begin{table}[!tbp]
\centering
\begin{tabular}{lc}
\toprule
\small{Architecture} & aIOU \% \\
\midrule
\small{Baseline (B)} & $66.99$ \\
\small{B + Class-Balanced Loss Weighting (C)} & $69.56$ \\
\small{B + C + Pose Auxiliary Task (P)} & $\mathbf{70.26}$ \\
\bottomrule
\end{tabular}
\caption{Performance for architectural additions to a single super-category  (`Large Animals') version of \textsc{SketchParse}.} 
\label{tab:1}
\end{table}

\noindent \textbf{Significance of class-balanced loss and auxiliary task:} To determine whether to incorporate class-balanced loss weighting and 2-D pose as auxiliary task, we conducted ablative experiments on a baseline version of \textsc{SketchParse} configured for a single super-category (`Large Animals'). As the results in Table \ref{tab:1} indicate, class-balanced loss weighting and inclusion of 2-D pose as auxiliary task contribute to improved performance over the baseline model. 

\begin{table*}[!tbp]
\centering
\begin{tabular}{cc}
\toprule
\small{Branch point/} & aIOU \% \\
\small{\# layers specialized for a super category} & \\
\midrule
\small{res5c / classifier block} & $58.94$ \\ 
\small{res5b / 1 res block + classifier block} & $\mathbf{60.19}$ \\ 
\small{res5a / 2 res blocks + classifier block} & $59.24$ \\ 
\small{res4b22 / 3 res blocks + classifier block} & $59.20$ \\
\small{res4b17 / 8 res blocks + classifier block} & $59.17$ \\
\bottomrule
\end{tabular}
\caption{Performance for various split locations within the scene parsing net~\cite{chen2016deeplab}. As suggested by the results, we use the layers upto  \textit{res5b} to instantiate the first level in \textsc{SketchParse}. The subsequent layers are used to instantiate expert super-category nets in the second level.}
\label{tab:splitpoint}
\end{table*} 

\noindent \textbf{Determining split point in base model:} A number of candidate split points exist which divide layers of the base scene parsing net~\cite{chen2016deeplab} into two disjoint groups. We experimented with different split points within the scene parsing net. For each split point, we trained a full $5$ super-category version of \textsc{SketchParse} model. Based on the results (see Table \ref{tab:splitpoint}), we used the split point (\textit{res5b}) which generated best performance for the final version viz. the \textsc{SketchParse} model with class-balanced loss weighting and pose estimation included for all super-categories. Note that we do not utilize the sketch router for determining the split point. From our experiments, we found the optimal split point results in shallow expert networks. This imparts \textsc{SketchParse} with better scalability. In other words, additional new categories and super-categories can be included without a large accompanying increase in number of parameters.

\begin{table*}[!tbp]
\resizebox{\textwidth}{!}{%
\centering
 \centering 
 \begin{tabular}{lccccccccccccc}
 \toprule 
      &                                & \multicolumn{2}{c}{Large Animals} &  \multicolumn{3}{c}{Small Animals} &  \multicolumn{2}{c}{4-Wheelers} & \multicolumn{2}{c}{2-Wheelers} &  \multicolumn{2}{c}{Flying Things} &  \\
 Model &  $\# parameters (M=millions)$ & \texttt{cow} & \texttt{horse} & \texttt{cat} & \texttt{dog} & \texttt{sheep} &  \texttt{bus} & \texttt{car} & \texttt{bicycle} & \texttt{motorbike} & \texttt{airplane} & \texttt{bird} & \textsc{Avg.} \\ 
 \midrule
 B-R5 & $88.4M$ & $66.52$ & $69.39$ & $62.62$ & $65.92$ & $67.65$ &  $54.42$  & $63.46$ & $59.27$  & $50.10$  &   $50.44$   &     $44.73$ & $\textcolor{red}{60.19}$\\
 BC-R5  & $88.4M$ &  $67.69$ & $70.02$ & $64.57$ & $68.54$ & $70.02$& $66.79$  & $66.96$ & $63.74$ & $53.15$ & $55.50$ & $47.72$ & $\textcolor{red}{62.90}$ \\   
 BCP-R-$1b$  & $65.4M$ &  $60.47$ & $22.09$ & $24.38$ & $22.76$& $23.87$& $19.96$  & $20.55$ & $18.21$ & $19.50$ & $21.94$ & $20.13$ & $\textcolor{red}{25.78}$ \\   
 BCP-R-$11b$  & $138.5 M$ &  $64.87$ & $65.69$ & $63.78$& $65.24$& $65.69$  & $64.91$ & $62.50$ & $57.44$ & $48.71$ & $51.47$ & $43.84$ & $\textcolor{red}{59.17}$ \\   
 \textbf{BCP-R5}  & $89M$ &  $68.10$ & $69.45$ & $65.47$ & $68.37$& $70.74$& $67.17$  & $66.48$ & $62.66$ & $54.49$ & $55.47$ & $48.77$ & $\textcolor{red}{\textbf{63.17}}$ \\   
 \textbf{BCP-R5 (100\% router)}  & $89M$ &  $68.78$ & $69.35$ & $69.60$ & $71.18$& $70.81$& $68.00$  & $67.35$ & $62.66$ & $55.04$ & $57.34$ & $50.89$ & $\textcolor{red}{\textbf{64.45}}$ \\   
 \bottomrule
 \end{tabular}
 }
\captionof{table}{Comparing the full $5$ super-category version of \textsc{SketchParse} (denoted \textbf{BCP-R5}) with baseline architectures.}
\label{table:3} 
\end{table*}

\noindent \textbf{Relative Loss weighting:} The `parsing' and `pose estimation' tasks are trained simultaneously. We weigh the losses individually and perform a grid search on the values of $\lambda$ and learning rate. Hence, total loss, $L_{tot}$ is

\begin{equation}
L_{tot} = L_{seg} + \lambda*L_{pose}
\end{equation}

The grid search is performed on the super category Large-Animals and the optimal $\lambda$ and learning rate value $(\lambda = 1$,  lr = $5 \times 10^{-4})$ (see Table \ref{tab:lambdaSearch}) are used for each branch in the final $5$ route network.

\begin{table}[!htbp]
\centering
\begin{tabular}{ccc}
\toprule
$\lambda$ & \small{learning rate} & aIOU \% \\
\midrule
0.1 & $1 \times 10^{-4}$ & $60.3$ \\
0.1 & $2.5 \times 10^{-4}$ & $61.17$ \\
0.1 & $5 \times 10^{-4}$ & $61.6$ \\
1 & $5 \times 10^{-4}$ & $\mathbf{63.50}$ \\
\bottomrule
\end{tabular}
\caption{Grid search over $\lambda$ and learning rate}
\label{tab:lambdaSearch}
\end{table}

\noindent \textbf{Full version net and baselines:} We compare the final $5$ super-category version (BCP-R5) of \textsc{SketchParse} (containing class-balanced loss weighting and pose estimation auxiliary task) with certain baseline variations -- (i) B-R5: No additional components included (ii) BC-R5: Class-balanced loss weighting included in B-R5 
(iii) BCP-R-$1b$: All categories are grouped into a single branch (iv) BCP-R-$11b$: A variant of the final version with a dedicated expert network for each category (i.e. one branch per category). 
 
 From the results (Table \ref{table:3}), we make the following observations: (1) Despite the challenges posed by hand-drawn sketches, our model performs reasonably well across a variety of categories (last but one row in Table \ref{table:3}). (2) Sketches from \textit{Large Animals} are parsed the best while those from \textit{Flying Things} do not perform as well. On closer scrutiny, we found that \texttt{bird} category elicited inconsistent sketch annotations given the relatively higher degree of abstraction in the corresponding sketches. (4) In addition to confirming the utility of class-balanced loss weighting and pose estimation, the baseline performances demonstrate that part (and parameter) sharing at category level is a crucial design choice, leading to better overall performance. In particular, note that having $1$ category per branch (BCP-R-$11b$) almost doubles the number of parameters, indicating poor category scalability. 
 
 To examine the effect of the router, we computed IOU by assuming  a completely accurate router for BCP-R5 (last row). This improves average IOU performance by 1.28. The small magnitude of improvement also shows that our router is quite reliable. The largest improvements are found among \textit{Small Animals} (`cat',`dog'). This is also supported by the router classifier’s confusion matrix (see Table \ref{tab:conf}). 
 
\begin{table}[!htbp]
\resizebox{\textwidth}{!}
{%
 \centering 
 \begin{tabular}{cccccccccccccc}
 \toprule 
            Directions & Large  & Small  &  4-Wheelers & 2-Wheelers & Flying &  AVG.\\
                       & Animals & Animals &  &  & Things &    \\       
 \midrule   
  8 &  $80.32$ & $53.92$ & $54.92$ & $59.89$ & $44.73$& ${\mathbf{58.45}}$ \\    
  4 &  $92.07$ & $76.78$ & $69.72$ & $87.16$ & $75.79$& ${\mathbf{80.44}}$ \\    
 \bottomrule
 \end{tabular}
 }
\captionof{table}{Performance of our best \textsc{SketchParse} model (\textbf{BCP-R5}) on the pose auxiliary task.}
\label{table:4} 
\end{table}

\begin{figure*}[!tbp]
    \includegraphics[width=\textwidth]{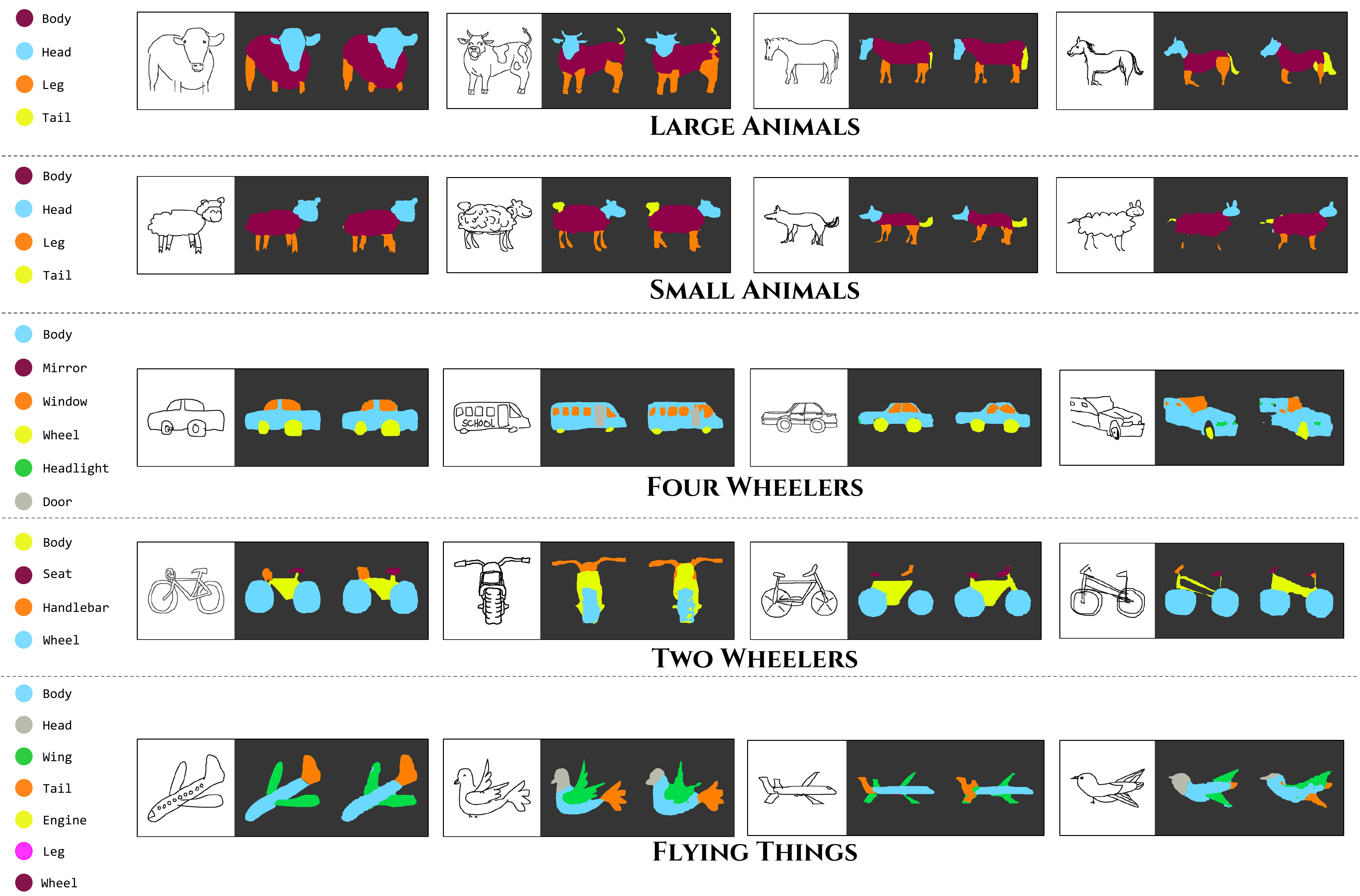}
    \caption{Qualitative part-parsing results across super-categories. Each row contains four panels of three images. In each panel, the test sketch is left-most, corresponding part-level ground-truth in the center and \textsc{SketchParse}'s output on the right. The four panels are chosen from the $100$th, $75$th, $50$th, $25$th percentile of each super-category's test sketches sorted by their average IOU. The background is color-coded black in all the cases.}
\label{fig:qualitative-sketches}
\end{figure*}

The performance of pose classifier can be viewed in Table \ref{table:4}. Note that simplifying the canonical pose directions (merging non-canonical directional labels with canonical directions) lends a dramatic improvement in accuracy. In addition, the most confusion among predictions is between the left-right directions and their corresponding perspective views (see Table \ref{tab:confPR5}). Depending on the granularity of pose information required we may merge the perspective directions as appropriate.

\begin{table*}[!htbp]
\footnotesize

\centering
\begin{tabular}{|c|c|c|c|c|}
\hline 
 & N & E & S & W \\
 \hline \hline
N & 0 & 6   & 5  & 3  \\ \hline
E & 0 & 329 & 24 & 47 \\ \hline
S & 0 & 12  & 55 & 15 \\ \hline
W & 0 & 62  & 19 & 410\\ \hline

\end{tabular}
\caption{Overall pose confusion matrix (4 directions)}
\label{tab:confPR5}
\end{table*}


\noindent \textbf{Qualitative evaluation:} Rather than cherry-pick results, we use a principled approach to obtain a qualitative perspective. We first sort the test sketches in each super-category by their aIOU (average IOU) values in decreasing order. We then select $4$ sketches located at $100$-th, $75$-th, $50$-th,  and $25$-th percentile in the sorted order. These sketches can be viewed in Figure \ref{fig:qualitative-sketches}. The part-level parsing results reinforce the observations made previously in the context of quantitative evaluation. 

\subsection{Parsing semantically related categories}
\label{sec:experiments-relatedcategories}

\begin{figure}[!ht]
\centering
 \includegraphics[width=\columnwidth]{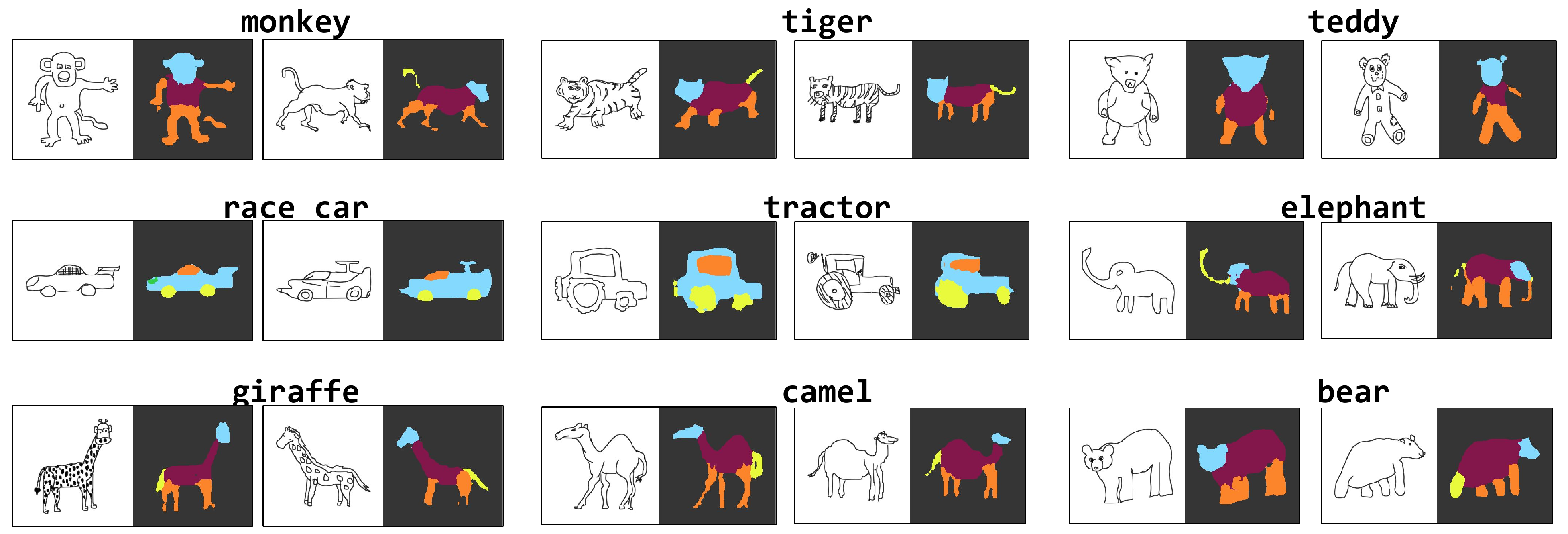}
 \caption{Part-parsing results for sketches from categories which are semantically similar to categories on which \textsc{SketchParse} is originally trained.}
\label{fig:sketchlab-semantically_related}
\end{figure}

We also examine the performance of our model for sketches belonging to categories our model is not trained on but happen to be semantically similar to at least one of the existing categories. Since segmentation information is unavailable for these sketches, we show two representative parsing outputs per class. We include the classes \texttt{monkey}, \texttt{tiger}, \texttt{teddy-bear}, \texttt{camel}, \texttt{bear}, \texttt{giraffe},  \texttt{elephant}, \texttt{race car} and \texttt{tractor} which are semantically similar to categories already considered in our formulation. As the results demonstrate  (Figure \ref{fig:sketchlab-semantically_related}), \textsc{SketchParse} accurately recognizes parts it has seen before (`head', `body', `leg' and `tail'). It also exhibits a best-guess behaviour to explain parts it is unaware of. For instance, it marks elephant `trunk' as either `legs' or `tails' which is a semantically reasonable error given the spatial location of the part. These experiments demonstrate the scability of our model in terms of category coverage. In other words, our architecture can integrate new, hitherto unseen categories without too much effort.

\subsection{Evaluating category-level scalability}
\label{sec:experiments-scalability}

As an additional experiment to evaluate performance scalability when categories are added incrementally, we performed the following experiment: (1) Train the BCP-R5 network with all categories, except the categories (`dog',`sheep') from \textit{Small Animals} super-category. (2) Freeze shared layers and fine-tune \textit{Small Animals} branch with `dog' data added. (3) Freeze shared layers and fine-tune the trained model from Step-(2) with `sheep' data added. 

\begin{table*}[!htbp]
\resizebox{\textwidth}{!}{%
\centering
 \centering 
 \begin{tabular}{lcccccccccccc}
 \toprule 
    & \multicolumn{2}{c}{Large Animals} &  \multicolumn{3}{c}{Small Animals} &  \multicolumn{2}{c}{4-Wheelers} & \multicolumn{2}{c}{2-Wheelers} &  \multicolumn{2}{c}{Flying Things} &  \\
 Model &  \texttt{cow} & \texttt{horse} & \texttt{cat} & \texttt{dog} & \texttt{sheep} &  \texttt{bus} & \texttt{car} & \texttt{bicycle} & \texttt{motorbike} & \texttt{airplane} & \texttt{bird} & \textsc{Avg.} \\ 
 \midrule
 Row-1            &  $67.20$ & $68.97$  & $65.67$ &         &         & $66.97$  & $66.13$ & $62.48$  & $51.23$ & $56.12$ & $50.35$ & $61.68$ \\
 Row-2            &  $66.87$ & $68.72$  & $64.14$ & $66.77$ &         & $65.86$  & $66.20$ & $62.48$  & $52.05$ & $56.37$ & $50.65$ & $62.01$ \\ 
 Row-3            &  $66.98$ & $68.68$  & $64.10$ & $65.03$ & $69.14$ & $66.29$  & $65.64$ & $62.77$  & $52.64$ & $56.10$ & $49.01$ & $62.18$ \\ 
 Row-4            &  $68.10$ & $69.45$  & $65.47$ & $68.37$ & $70.74$ & $67.17$  & $66.48$ & $62.66$  & $54.49$ & $55.47$ & $48.77$ & $63.17$ \\   
 \bottomrule
 \end{tabular}
 }
\captionof{table}{Comparing the full $5$ super-category version of \textsc{SketchParse} (denoted \textbf{BCP-R5}) with baseline architectures (Row-1: Model trained without `sheep',`dog' categories, Row-2: Previous model (Row-1's) fine-tuned with only `dog' data added, Row-3: Fine-tune previous (i.e. `dog' fine-tuned) model with `sheep' data added, Row-4: Original result (where all categories are present from the beginning -- same as last but one row of Table \ref{table:3})).
}
\label{table:scalability} 
\end{table*}


We observe (Table \ref{table:scalability}) that the average IOU (last column) progressively improves as additional categories are added. In other words, overall performance does not drop even though shared layers are frozen. This shows the scalable nature of our architecture. Of course, the IOU values are slightly smaller compared to the original result (last row) where all categories are present from the beginning, but that is a consequence of freezing shared layers. 

\subsection{Fine-grained retrieval}
\label{sec:experiments-fgr}

\begin{figure}[!htbp]
\centering
     \includegraphics[width=\textwidth,height=0.9\textheight]{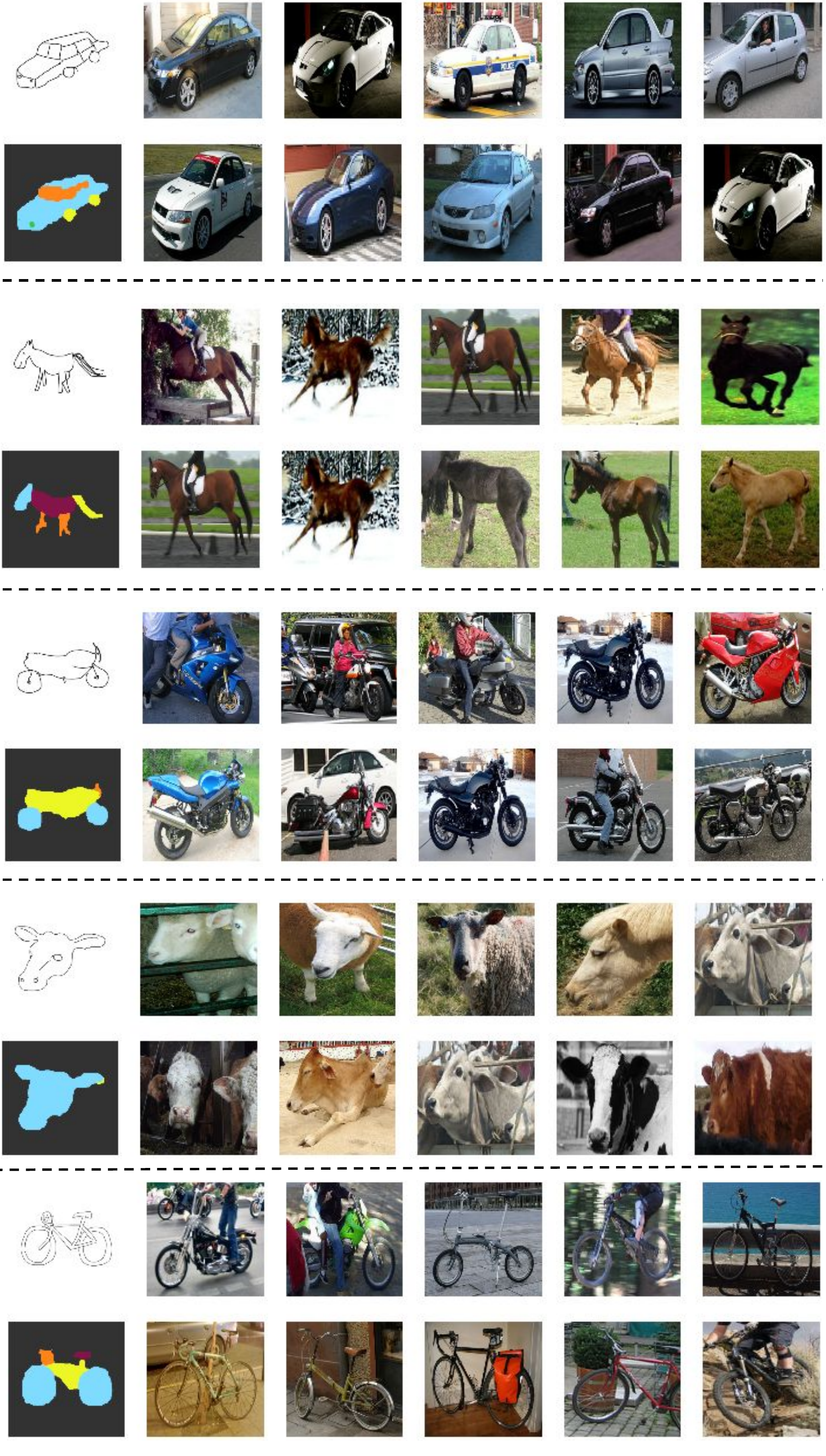}
    \caption{Five sketch-based image retrieval panels are shown. In each panel, the top-left figure is the query sketch. Its part parsing is located immediately below. Each panel has two sets of retrieval results for the presented sketch. The first row corresponds to Sangkloy et al.'s~\cite{Sangkloy:2016:SDL:2897824.2925954} retrieval results and the second contains our re-ranked retrievals based on part-graphs (Section \ref{sec:experiments-fgr}).}
\label{fig:retrieval}
\end{figure}

In another experiment, we determine whether part-level parsing of sketches can improve performance for existing sketch-based image retrieval approaches which use global image-level features. We use the PASCAL parts dataset~\cite{chen_cvpr14}, consisting of $9620$ photo images across $10$ categories, as the retrieval database. As a starting point, we consider the sketch-based image retrieval system of Sangkloy et al.~\cite{Sangkloy:2016:SDL:2897824.2925954}. The system consists of a trained Siamese Network model which projects both sketches and photos into a shared latent space. We begin by supplying the query sketch $I$ from our dataset~\cite{eitz2012humans} and obtain the sequence of retrieved PASCAL parts images $D_1,D_2,\ldots D_T$. Suppose the part-segmented version of $I$ is $I_p$. We use a customized Attribute-Graph approach~\cite{prabhu2015attribute} and construct a graph $G_I$ from $I_p$. The attribute graph is designed to capture the spatial and semantic aspects of the part-level information at local and global scales. We use annotations from the PASCAL parts dataset to obtain part-segmented versions of retrieved images, which in turn are used to construct corresponding attribute graphs $G_{D_1},G_{D_2},\ldots G_{D_T}$.

Each graph has two kinds of nodes: a single global node and a local node for each non-contiguous instance of a part present in the segmentation output from \textsc{SketchParse}. The global node attributes include:
\begin{itemize}
    \item a histogram that keeps a count of each type of part present in the image
    \item the non-background area in the image as a fraction of total area
\end{itemize}
A local node is instantiated at every non-contiguous part present in the segmentation output. We drop nodes for which the corresponding part area is less than 0.1\% of the total non-background area with the assumption that these are artifacts in the segmentation output. Each local node has the following attributes: 
\begin{itemize}
    \item angle subtended by the part at the centre of the sketch
    \item centre of the part
\end{itemize}
\noindent Edges are present between local nodes corresponding to parts that have a common boundary. Each such edge encodes the relative position of both parts using a polar coordinate system.

Every local node is also connected to the global node. These edges encode the absolute position of the part in the image. Again, the area of each part is used as a multiplicative weight for each similarity computation it's corresponding node participates in.

For re-ranking the retrieved images, we use Reweighted Random Walks Graph Matching~\cite{cho2010reweighted} to compute similarity scores between $G_I$ and $G_{D_i}, 1 \leqslant i \leqslant T$, although any other graph matching algorithm that allows incorporating constraints may be used. During the graph matching process we enforce two constraints:
\begin{itemize}
    \item A global node can only be matched to a global node of the other graph.
    \item Local nodes can only be matched if they correspond to the same type of part (\textit{e.g}. local nodes corresponding to legs can only be matched to other legs and cannot be matched to other body parts) 
\end{itemize}

For our experiments, we examine our re-ranking procedure for top-$50$ (out of $9620$ images) of Sketchy model's retrieval results. In Figure \ref{fig:retrieval}, each panel corresponds to top-$5$ retrieval results for a particular sketch. The sketch and its parsing are displayed alongside the $5$ nearest neighbors in latent space of Sketchy model (top row) and the top $5$ re-ranked retrievals using our part-graphs (bottom row). The results show that our formulation exploits the category, part-level parsing and pose information to obtain an improved ranking. 

\subsection{Describing sketches in detail}
\label{sec:experiments-det}

\begin{figure}[!tbp]
\centering
\includegraphics[width=\textwidth,keepaspectratio]{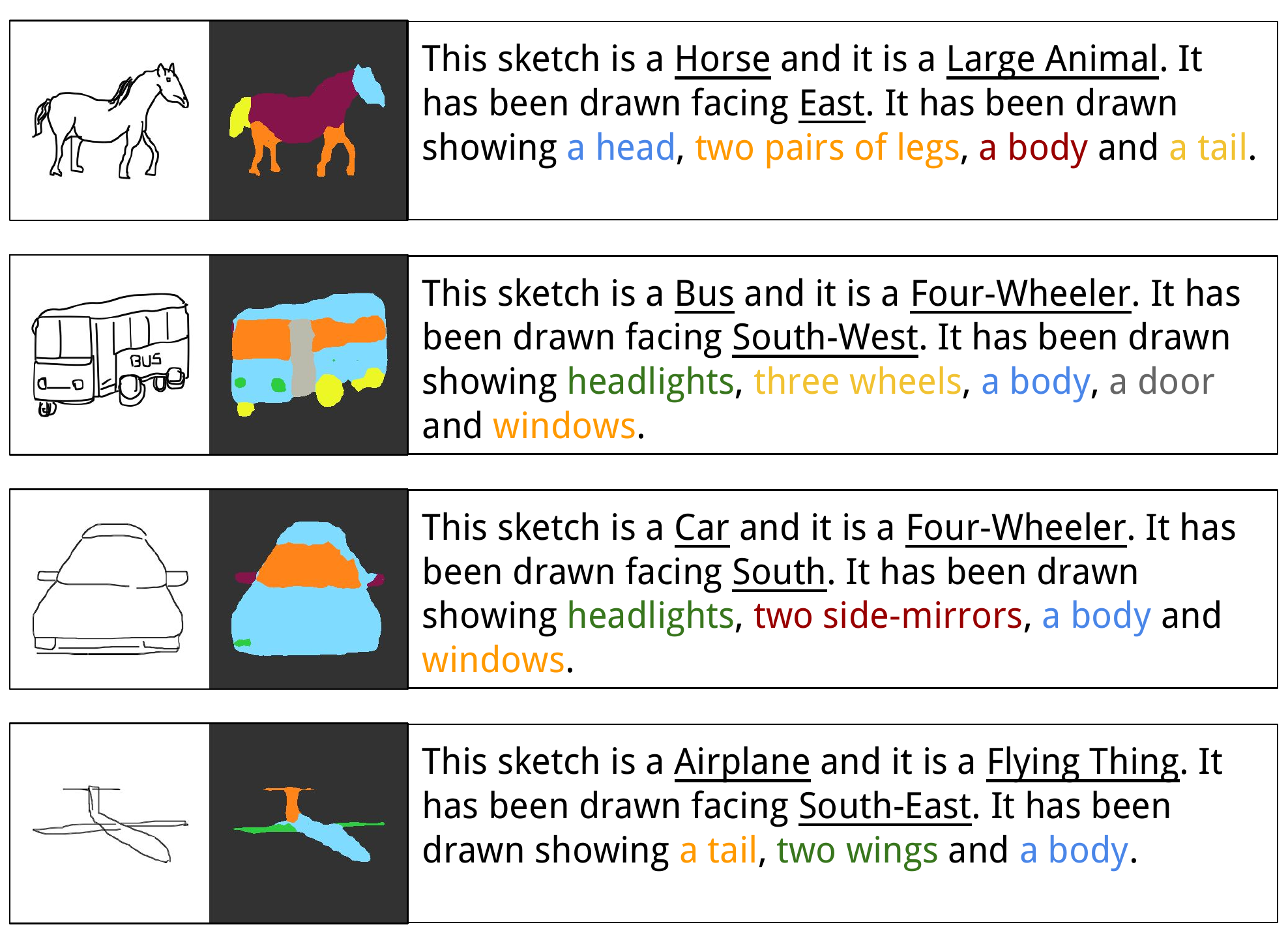}
\caption{Some examples of fine-grained sketch descriptions. Each panel above shows a test sketch (left), corresponding part-parsing (center) and the description (last column). Note that in addition to parsing output, we also use the outputs of auxiliary pose network and router classifier to generate the description. The color-coding of part-name related information in the description aligns with the part color-coding in the parsing output. See Section \ref{sec:experiments-det} for additional details.}
\label{fig:fine-grained-description}
\end{figure}

Armed with the information provided by our model, we can go beyond describing a hand-drawn sketch by a single category label. For a given sketch, our model automatically provides its category, associated super-category, part-labels and their counts and 2-D pose information. From this information, we use a template-filling approach to generate descriptions -- examples can be seen alongside our qualitative results in Figure \ref{fig:fine-grained-description}. A fascinating application, inspired by the work of Zhang et al.~\cite{zhang2016stackgan}, would be to use such descriptions to generate freehand sketches using a Generative Adversarial Network approach. 

\section{Conclusion}
\label{sec:conclusion}

Given the generally poor drawing skills of humans and sparsity of detail, it is very challenging to simultaneously recognize and parse sketches across multiple groups of categories. In this paper, we have presented \textsc{SketchParse}, the first deep-network architecture for fully automatic parsing of freehand object sketches. The originality of our approach lies in successfully repurposing a photo scene-segmentation net into a category-hierarchical sketch object-parsing architecture. The general nature of our transfer-learning approach also allows us to leverage advances in fully convolutional network-based scene parsing approaches, thus continuously improving performance. Another novelty lies in obtaining labelled training data \textit{for free} by sketchifying photos from object-part datasets, thus bypassing burdensome annotation step. Our work stands out from existing approaches in the complexity of sketches, number of categories considered and semantic variety in categories. While existing works focus on one or two super-categories and build separate models for each, our scalable architecture can handle a larger number of super-categories, all with a single, unified model. Finally, the utility of \textsc{SketchParse}'s novel multi-task architecture is underscored by its ability to enable applications such as fine-grained sketch description and improving sketch-based image retrieval. 

Please visit \url{https://github.com/val-iisc/sketch-parse} for pre-trained models, code and resources related to the work presented in this paper.

For future exploration, it would be interesting to explore additional auxiliary tasks such as adversarial loss~\cite{luc2016semantic} and part-histogram loss~\cite{liang2015proposal} to further boost part-parsing performance. Another natural direction to pursue would be the viability of \textsc{SketchParse}'s architecture for semantic parsing of photo objects. 

\bibliographystyle{ieee}

\section*{Confusion Matrices for Pose}
\label{sec:app}

\begin{table*}[!htbp]

\footnotesize
\centering
\begin{tabular}{|c|c|c|c|c|c|c|c|c|}
\hline 
 & N & NE & E & SE & S & SW & W & NW \\
 \hline \hline
N & 0 & 0 & 0  & 2 & 0 & 0 & 0  & 0 \\ \hline
NE& 0 & 0 & 1  & 1 & 0 & 0 & 0  & 0 \\ \hline
E & 0 & 0 & 50 & 2 & 0 & 0 & 2  & 0 \\ \hline
SE& 0 & 0 & 3  & 6 & 0 & 0 & 0  & 0 \\ \hline
S & 0 & 0 & 1  & 1 & 7 & 4 & 0  & 0 \\ \hline
SW& 0 & 0 & 0  & 0 & 3 & 9 & 5  & 0 \\ \hline
W & 0 & 0 & 0  & 0 & 2 & 9 & 79 & 0 \\ \hline
NW& 0 & 0 & 0  & 0 & 0 & 1 & 0  & 0 \\ \hline
\end{tabular}
\caption{Pose confusion matrix for Large Animals}
\label{tab:confPR1}
\end{table*}

\begin{table*}[!htbp]

\footnotesize
\centering
\begin{tabular}{|c|c|c|c|c|c|c|c|c|}
\hline 
 & N & NE & E & SE & S & SW & W & NW \\
 \hline \hline
N & 0 & 0 & 0  & 0  & 1  & 0  & 1  & 0 \\ \hline
NE& 0 & 0 & 5  & 4  & 3  & 0  & 0  & 0 \\ \hline
E & 0 & 0 & 30 & 12 & 7  & 1  & 4  & 0 \\ \hline
SE& 0 & 0 & 5  & 8  & 10 & 1  & 0  & 1 \\ \hline
S & 0 & 0 & 2  & 2  & 44 & 6  & 3  & 0 \\ \hline
SW& 0 & 0 & 2  & 0  & 2  & 5  & 10 & 0 \\ \hline
W & 0 & 0 & 4  & 3  & 8  & 20 & 64 & 0 \\ \hline
NW& 0 & 0 & 1  & 2  & 1  & 0  & 8  & 0 \\ \hline
\end{tabular}
\caption{Pose confusion matrix for Small Animals}
\label{tab:confPR2}
\end{table*}

\begin{table*}[!htbp]
\footnotesize
\centering
\begin{tabular}{|c|c|c|c|c|c|c|c|c|}
\hline 
 & N & NE & E & SE & S & SW & W & NW \\
 \hline \hline
N & 0 & 0 & 0  & 1  & 1 & 0  & 0  & 0 \\ \hline
NE& 0 & 0 & 0  & 0  & 0 & 2  & 1  & 0 \\ \hline
E & 0 & 2 & 27 & 7  & 0 & 1  & 16 & 0 \\ \hline
SE& 0 & 1 & 2  & 14 & 1 & 0  & 0  & 0 \\ \hline
S & 0 & 0 & 0  & 0  & 2 & 0  & 1  & 0 \\ \hline
SW& 0 & 0 & 0  & 0  & 0 & 12 & 2  & 0 \\ \hline
W & 0 & 1 & 14 & 3  & 0 & 7  & 23 & 0 \\ \hline
NW& 0 & 0 & 0  & 1  & 0 & 0  & 0  & 0 \\ \hline
\end{tabular}
\caption{Pose confusion matrix for 4 Wheelers}
\label{tab:confPR3}
\end{table*}

\begin{table*}[!htbp]
\footnotesize
\centering
\begin{tabular}{|c|c|c|c|c|c|c|c|c|}
\hline 
 & N & NE & E & SE & S & SW & W & NW \\
 \hline \hline
N & 0 & 0 & 0  & 0  & 0  & 0  & 1  & 0 \\ \hline
NE& 0 & 0 & 8  & 0  & 0  & 0  & 0  & 0 \\ \hline
E & 0 & 3 & 57 & 1  & 0  & 1  & 8  & 0 \\ \hline
SE& 0 & 0 & 5  & 5  & 0  & 0  & 2  & 0 \\ \hline
S & 0 & 0 & 0  & 1  & 0  & 1  & 0  & 0 \\ \hline
SW& 0 & 0 & 0  & 0  & 0  & 6  & 5  & 0 \\ \hline
W & 0 & 0 & 11 & 0  & 0  & 21 & 43 & 3 \\ \hline
NW& 0 & 0 & 0  & 0  & 0  & 3  & 2  & 1 \\ \hline
\end{tabular}
\caption{Pose confusion matrix for 2 Wheelers}
\label{tab:confPR4}
\end{table*}

\begin{table*}[!htbp]
\footnotesize
\centering
\begin{tabular}{|c|c|c|c|c|c|c|c|c|}
\hline 
 & N & NE & E & SE & S & SW & W & NW \\
 \hline \hline
N & 0 & 0 & 2  & 1 & 3 & 1  & 1  & 0 \\ \hline
NE& 0 & 0 & 10 & 5 & 0 & 1  & 2  & 0 \\ \hline
E & 0 & 1 & 41 & 5 & 2 & 1  & 1  & 0 \\ \hline
SE& 0 & 0 & 3  & 5 & 1 & 2  & 0  & 0 \\ \hline
S & 0 & 0 & 3  & 2 & 2 & 0  & 0  & 0 \\ \hline
SW& 0 & 0 & 2  & 1 & 0 & 6  & 5  & 0 \\ \hline
W & 0 & 0 & 5  & 8 & 1 & 25 & 31 & 0 \\ \hline
NW& 0 & 0 & 2  & 2 & 2 & 3  & 2  &  \\ \hline
\end{tabular}
\caption{Pose confusion matrix for Flying Things}
\label{tab:confPR5}
\end{table*}

\begin{table*}[!htbp]
\footnotesize
\centering
\begin{tabular}{|c|c|c|c|c|c|c|c|c|}
\hline 
 & N & NE & E & SE & S & SW & W & NW \\
 \hline \hline
N & 0 & 0 & 2   & 4  & 5  & 1  & 2   & 0 \\ \hline
NE& 0 & 0 & 24  & 10 & 3  & 3  & 3   & 0 \\ \hline
E & 0 & 6 & 205 & 27 & 9  & 4  & 31  & 0 \\ \hline
SE& 0 & 1 & 18  & 38 & 12 & 3  & 2   & 1 \\ \hline
S & 0 & 0 & 6   & 6  & 55 & 11 & 4   & 0 \\ \hline
SW& 0 & 0 & 4   & 1  & 5  & 38 & 27  & 0 \\ \hline
W & 0 & 1 & 34  & 14 & 11 & 82 & 240 & 3 \\ \hline
NW& 0 & 0 & 3   & 5  & 3  & 7  & 12  & 1 \\ \hline
\end{tabular}
\caption{Overall pose confusion matrix}
\label{tab:confPR5}
\end{table*}

\end{document}